\RequirePackage{fix-cm}

\documentclass[smallextended]{svjour3}       
\smartqed  
\usepackage[round]{natbib} 

\usepackage{graphicx}
\usepackage{amsmath}
\usepackage{amsfonts}
\usepackage[utf8]{inputenc}
\usepackage[table]{xcolor}
\usepackage{arydshln} 
\usepackage{pgfplotstable} 
\usepackage{capt-of}
\usepackage{xcolor}
\usepackage{algorithm, algpseudocode}
\usepackage{dsfont} 

\journalname{Data Mining and Knowledge Discovery}

\newcommand{\aucw}{\ensuremath{\textrm{AUC}_w}}
\newcommand{\auca}{\ensuremath{\textrm{AUC@}\alpha}}
\newcommand{\tpra}{\ensuremath{\textrm{TPR@}\alpha}}
\newcommand{\preca}{\ensuremath{\textrm{precision@}p}}
\newcommand{\fa}{\ensuremath{\textrm{F1@}\alpha}}
\newcommand{\vola}{\ensuremath{\textrm{VOL@}\alpha}}
\newcommand{\cvola}{\ensuremath{\textrm{CVOL@}\alpha}}

\begin{document}

\title{Is AUC the best measure for practical comparison of anomaly detectors?}

\author{V\'it \v{S}kv\'ara      \and
    Tom\'a\v{s} Pevn\'y      \and
        V\'aclav \v{S}m\'idl
}

\institute{V\'it \v{S}kv\'ara$^1$ \at
              \email{skvara@utia.cas.cz}
           \and
              Tom\'a\v{s} Pevn\'y$^2$ \at
              \email{pevnytom@fel.cvut.cz}
           \and
           V. \v{S}m\'idl$^1$ \at
              \email{smidl@utia.cas.cz}           
           \and \at
              $^1$UTIA CAS CR, Pod Vod\'arenskou v\v{e}\v{z}\'i 4, Prague, Czech Republic, 18200 \\
              $^2$FEL CTU Prague, Technick\'a 2, Prague, Czech Republic, 16000
}


\date{Received: date / Accepted: date}

\maketitle

\begin{abstract}
The area under receiver operating characteristics (AUC) is the standard measure for comparison of anomaly detectors. Its advantage is in providing a scalar number that allows a natural ordering and is independent on a threshold, which allows to postpone the choice. In this work, we question whether AUC is a good metric for anomaly detection, or if it gives a false sense of comfort, due to relying on assumptions which are unlikely to hold in practice. Our investigation shows that variations of AUC emphasizing accuracy at low false positive rate seem to be better correlated with the needs of practitioners, but also that we can compare anomaly detectors only in the case when we have representative examples of anomalous samples. This last result is disturbing, as it suggests that in many cases, we should do active or few-show learning instead of pure anomaly detection.
\end{abstract}

\section{Motivation}
The goal of anomaly detection is to find samples occurring with such a small probability that they seem to be generated by some other process. While this definition makes sense from the point of view of probability theory, the practitioners are interested only in anomalies of a certain kind. For example, the use of anomaly detection in computer security is motivated by the assumption that attacks are rare, but not every rare event is an attack.

Anomaly detection has a long history and there are hundreds of models based on vastly different approaches, such as modifications of the k--nearest neighbors algorithm \citep{harmeling2006outliers}, random forests \citep{liu2008isolation}, gaussian mixture models \citep{mahadevan2010anomaly}, kernel density estimates \citep{latecki2007outlier}, histogram--based models \citep{pevny2016loda} or neural networks \citep{schlegl2017unsupervised}. Several comparative studies exist -- e.g. \citep{goldstein2016comparative, pimentel2014review,lazarevic2003comparative, chandola2009anomaly} or \citep{markou2003novelty} -- in which the authors present an overview of existing anomaly detection methods, sometimes with a direct comparison on benchmark datasets. However, the authors never ask the question if the way the methods are compared to each other is appropriate for the anomaly detection setting. The area under the receiver operating characteristic curve (ROC AUC or just AUC in this text) seems to be the gold-standard of the field, as it allows us to compare models by a single number. In \citep{vanderlooy2008critical}, the authors compare "soft" variants of AUC in which the contribution of a sample to the total value of AUC is weighted by the degree in which it is labeled correctly or incorrectly. They demonstrate that none of these alternatives can systematically outperform the conventional AUC, however they do so on a set of binary classification experiments, without any consideration for the specifics of anomaly detection.

\begin{figure}
\centering
  \includegraphics[width=0.6\columnwidth]{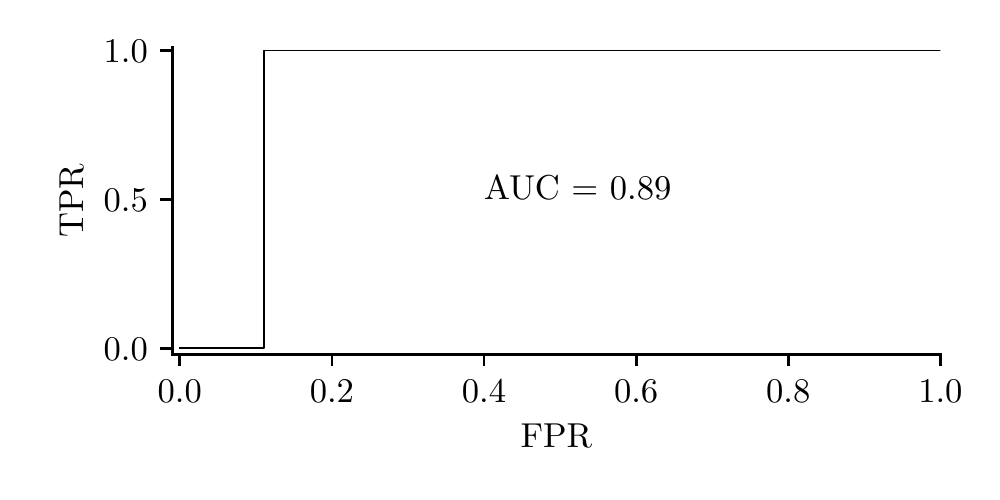}
  \caption{Receiver operating characteristic (ROC) curve and the corresponding AUC of a degenerate anomaly detector. FPR / TPR stands for false / true positive rate.}
  \label{fig:bad_auc}
\end{figure}

While AUC is a darling performance measure of anomaly detection of researchers, practitioners rely on other measures better reflecting their needs. For example in intrusion detection, precision@k is popular \citep{grill2016learning}, since the security officer can investigate at most $k$-most anomalous samples a day and he wishes to find as many dangerous security incidents present in the data as possible. Similarly, the true positive rate at a fixed false positive rate is used in tuning anti-virus engines, as these applications need to be certain not to raise too many alarms to the users (discussions with experts from the field suggests the tolerable false positive rate to be in order $10^{-4}$ and below). Other security application domains, fraud detection, control of industrial and environmental processes, and others have similar application constraints emphasizing low false positive rates. We believe not to exaggerate if we state that for the purpose of anomaly detection, the region of low false positive rates in the ROC curve might be the most critical for evaluating the performance of a method. 

Imagine now a data-scientist developing new anomaly detection tools supporting the above applications. They wish to implement a state of the art algorithm, which can represent a significant time investment, yet all papers available to them compare the proposed detectors using AUC. How does AUC help them to decide which method to implement if he is interested in performance measured by precision@k? AUC may be misleading since it is an integral over all possible thresholds, putting equal emphasis on each threshold, where most of them are far off his point of interest. How much are they certain that the proposed detector is not degenerated, as is shown in Fig.~\ref{fig:bad_auc}?

We admit that other measures than AUC are sometimes reported. In \citep{mahadevan2010anomaly}, equal error rate -- the percentage of misclassified frames when the false positive rate is equal to the false negative rate -- is reported together with AUC. In the anomaly detection survey \citep{campos2016evaluation}, average precision and precision@k adjusted for class imbalance are reported together with AUC and it is concluded that their behavior is very dataset-dependent and less stable than that of AUC, but no deeper insight is given.

In this paper, the subject of interest are performance measures instead of anomaly detection algorithms themselves. Our goal is to shed more light on the behavior of different measures. Particularly we want to quantify how different measures are correlated with each other and what happens if a detector is selected using a different measure than the one that is used in the application. We endeavor to suggest an alternative measure which would be more descriptive of the performance for practitioners.         

While the problem of choosing the right measure seems to be solvable using some variants of AUC, we also demonstrate a more profound problem: if the validation set contains examples of anomalies with different statistical properties (probability distributions) than anomalies in the testing set, the performance measures are not informative. The only exception is the volume of the decision region, which does not rely on anomalous samples and is theoretically justified \citep{steinwart2005classification}. Unfortunately, the volume of the decision region is difficult to estimate in higher dimensions. 

The rest of the paper is organized as follows: the next sections contain the description of tested measures, datasets and algorithms. Afterward, a description of the undertaken experiments is given together with their evaluation. Results of these experiments and their implications are discussed at the end of the paper together with concluding remarks.

\section{Performance measures}
\label{sec:measures}
Anomaly detection problems can be divided into categories based on the nature and availability of the data and the methods that are suitable for their solving. Following the categorization that is outlined e.g. in \citep{hodge2004survey}, the categories are:

\begin{description}
  \item \textbf{Supervised} anomaly detection where examples from both anomalous and normal classes are available for training. Such tasks are tackled similarly to binary classification and a representation of both classes is learned.
  \item \textbf{Semisupervised} anomaly detection where examples from both classes are also known, however only the normal class is used and learned during training and anomalies are used for validation. This approach is sometimes called \textit{one-class} classification or \textit{novelty detection} \citep{pimentel2014review}. The reason for learning only the normal class is usually that the anomalies are very sparse and so diverse that they cannot be considered to represent a second class.
  \item \textbf{Unsupervised} anomaly detection where the methods try to determine the anomalies with no prior knowledge of the data. The dataset is evaluated as a whole, a model of the normal class is established and the most non-normal points are reported as anomalies. This approach assumes that anomalies are far less frequent. Semisupervised techniques can be adapted for the unsupervised setting when a sample of unlabeled data is used as the training dataset, under the assumption that the model is robust to the possible contamination -- presence of anomalous data.
\end{description}

The original scenario in which the receiver operating characteristic (ROC) curve \citep{egan1975signal} and the corresponding area under the curve (AUC) was used as a performance measure is binary classification, which is closest to supervised anomaly detection setting. However, it is commonly used for evaluation in all of the three separate settings without any hesitation, even though there is no real second class in the semi- and unsupervised cases. The typical argument justifying its use is that the precise application conditions enabling to measure detection accuracy at a precise false positive rate are unknown, therefore AUC offers a good solution as it summarizes all application conditions. While this is true, it also means that these other application conditions are very different from those of our interest, which possibly negatively influences our decision. 

Below, the performance measures studied in this paper are listed. The list is far from complete, but it lists those that are of interest to most practitioners and also those that are interesting from a theoretical point of view. 

We assume that for an input $x$ an anomaly detection algorithm produces a real-valued anomaly score instead of a binary value, i.e. it is a projection $f:\mathcal{X} \rightarrow \mathbb{R}$ from the sample space $\mathcal{X}$. We also assume that a higher score corresponds to samples more likely to be anomalous. Anomaly score does not have to be a probability in the range $[0, 1]$ or even a positive number (e.g. in the OC-SVM model). 

When a sample is to be labeled as normal/anomalous, the output of the detector is compared to a threshold. Its value is typically determined on basis of tolerated false positive rate and an estimate of the true contamination rate of a dataset $X$, which we define as $C(X)=\frac{P}{P+N}$, where $P$ and $N$ is the total number of positive and negative samples in $X$.

\begin{table}
\centering
	\begin{tabular}{c | c c}
		true label/estimated label & normal & anomalous \\
		\hline
		normal & tn & fp  \\
		anomalous & fn & tp 
	\end{tabular}
	\caption{A confusion matrix of a model.}
	\label{tab:conf_ex}
\end{table}
Table~\ref{tab:conf_ex} displays the confusion table that introduces basic concepts and notation needed below. It summarizes the performance of an algorithm with a particular threshold by presenting the total number of correctly (tp = true positives and tn = true negative) and incorrectly (fp = false positives and fn = false negatives) identified samples. 

A public repository containing the implementation of all measures used in this paper using the Julia language \citep{bezanson2012julia} is available at \texttt{https://github.com/vitskvara/EvalCurves.jl}.

\subsection{Area under the ROC curve}
The most widely used measure in the field of anomaly detection is the area under the ROC curve (the acronym AUC will be used in the following text for the sake of brevity). The ROC curve is a parametric curve describing the trade-off between true positive rate $\text{TPR}(\tau) = \frac{\text{tp}}{\text{tp+fn}}(\tau)$ and false positive rate $\text{FPR}(\tau) = \frac{\text{fp}}{\text{fp+tn}}(\tau)$ for different values of the decision threshold $\tau$. A ROC curve can be easily computed on a dataset from the knowledge of the true labels and anomaly scores of the samples. 

\begin{figure}
\centering
\includegraphics[scale=0.85]{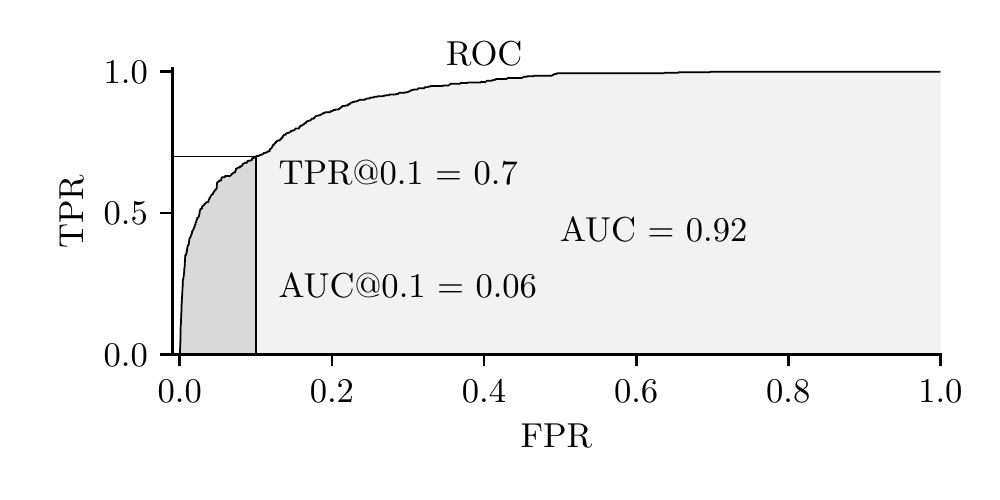}
\caption{An example of ROC curve and the derived measures based on FPR=0.1. AUC is the whole shaded area under the ROC curve. The darker shading corresponds to AUC@0.1.}
\label{fig:ROC}
\end{figure}

Then, the area under the curve is calculated as the following integral
\begin{equation}
\label{eq:auc}
\text{AUC}=\int_{\mathbb{R}}\text{TPR}(\tau)d\text{FPR}^{\prime}(\tau)d\tau = \int_0^1\text{TPR}(\text{FPR})d\text{FPR}.
\end{equation}
The last integral that uses $\text{TPR}(\cdot)$ as a function of the corresponding FPR shows the simple concept behind the AUC that can be easily discerned from a ROC curve drawn in a graph. An example of a ROC curve and the corresponding AUC is in Fig.~\ref{fig:ROC}. In practice, the corresponding AUC is estimated from an empirical ROC curve using some numerical integration scheme, e.g. the trapezoidal rule.

As mentioned above, the main advantage of AUC is that it does not depend on the choice of a particular decision threshold. Also, the measure has a straightforward interpretation -- it is an estimate of the probability that a randomly chosen positive sample is ranked higher than a randomly chosen negative sample \citep{hand2001simple}. However, a lot of information is lost when the whole ROC curve is summarized into a single number. This is especially concerning for the case of anomaly detection, where usually the region of low false positive rates is of interest, since anomalies are sparse compared to normal data and we strive to achieve a low false positive rate. It is frequent in security applications to draw ROC curve with a logarithmic scale on the x-axis.

\subsection{\auca}
A simple alternative to AUC is calculated only up to some value of false positive rate $\alpha$. Numerically, it is important to interpolate the ROC curve for a given $\alpha$ before computing the integral, especially for datasets with a small number of samples. In Fig.~\ref{fig:ROC}, AUC@0.1 corresponds to the darker grey region. \auca\ can be easily normalized by dividing by the chosen $\alpha$, in which case the best detector has $\auca = 1$ similarly to AUC.

\subsection{\tpra}
Another performance measure popular among practitioners (anti-virus engine vendors) is simply the true positive rate (TPR) evaluated at a given false positive rate (FPR) $\alpha.$ This measure can be easily read from a ROC curve and in practice, as in the case of \auca,  it is necessary to interpolate the ROC curve since FPR has discrete values.

\subsection{Weighted AUC}
\tpra~belongs to a class of Neyman Pearson measures \citep{scott2007performance}, where the goal is to find a classifier minimizing error of one type while the error of the other is bound by some value. It has been shown in \citep{scott2007performance}, that for a given bound on the false positive rate $\alpha$, the best classifier $f$ from a class of classifiers $\mathcal{F}$ consistently minimizing the true positive rate can be found as
\begin{equation}
\arg \min_{f\in\mathcal{F}} \frac{1}{\alpha} \max\{\text{FPR}(f) - \alpha, 0\} + (1 - \text{TPR}(f)),
\end{equation}   
where $\text{FPR}(\cdot)$ and $\text{TPR}(\cdot)$ denotes the false and true positive rate of a particular classifier.

This puts more weight on the region of low FPR values proportionate to $1/\text{FPR}(\alpha)$. In a manner similar to the definition of AUC~\eqref{eq:auc}, we can define the weighted AUC as an integral over all values $\alpha$ as
\begin{equation}
\aucw=\int_0^1\text{TPR}(\text{FPR})\text{FPR}^{-1}d\text{FPR},
\label{eq:AUCw}
\end{equation}
where we define $\text{FPR} = 0$ as $\text{TPR}(\text{FPR})\text{FPR}^{-1} = 0$, which makes the integral proper. While it might seem that this definition may still result in an arbitrary large number, in practice, the FPR in the ROC curve changes by a given finite increment that equals $1/p$, where $p$ is the total number of positive samples. This means that \aucw~attains comparable values, although it is not easily normalized like the remaining measures described here, which might create some issues in the former analysis.

\begin{figure}
\centering
\includegraphics[scale=0.85]{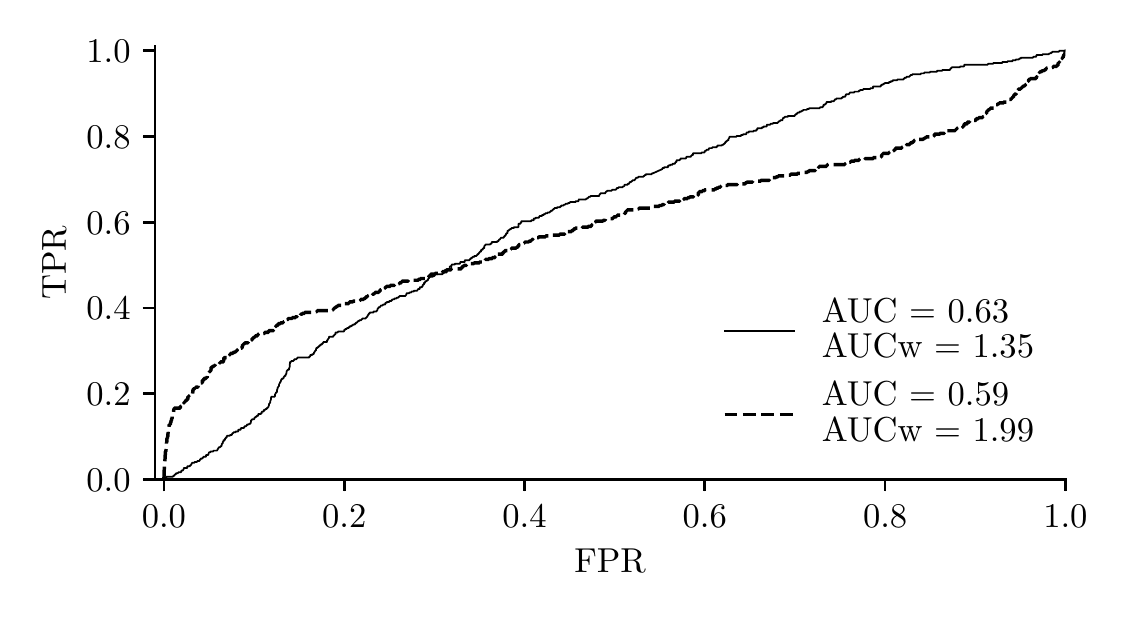}
\caption{Two ROC curves with different AUC and \aucw.}
\label{fig:AUCw}
\end{figure}

Fig.~\ref{fig:AUCw} illustrates the motivation behind \aucw~showing ROC curves of two detectors. The first detector with the ROC curve drawn with solid has a higher AUC than the second detector with ROC curve drawn in dashed line. If we base the selection on AUC, we would choose the first detector. Yet, this detector has inferior performance in the area of low false positive rate, where the second is better. If the decision would be based on the \aucw, the second detector would be selected. 

\subsection{\preca} \label{preca}
In binary classification, the precision, given a real--valued threshold $\tau$, is defined as $\text{PREC}(\tau)=\frac{\text{tp}}{\text{tp + fp}}(\tau)$. To make this measure more appropriate for anomaly detection setting, only $p\%$ most anomalous samples can be taken into account. Unfortunately, this measure is difficult to compare across datasets, because it depends on the proportion of anomalies with respect to normal samples. To make the measure comparable, the definition has been adapted as follows. First, anomalies are randomly removed from the testing dataset on which we want to evaluate the detector such that their proportion is $p$\%. This sub-sampled testing dataset is denoted $X$, and the \preca~is calculated as follows
\begin{equation} \label{eq:preca}
  \preca = \frac{|\lbrace x \in X_p \land \text{ true label of } x \text{ is positive}  \rbrace|}{|X_p|},
\end{equation}
where $X_p$ is the set of $p\%$ most anomalous samples in $X$. With a perfect detector, $X_p$ should contain all the anomalies and nothing else, thus yielding \preca = 1. Note that the random sampling may leave very distinct anomalies in $X$. This leads to very noisy values of \preca for small datasets. Therefore, it is reasonable to randomly sample multiple sets $X$ and then compute an average \preca\ on those.

\subsection{\fa} \label{fa}
\textbf{F1--score} is a single number computed as $\text{F}1(\tau) = \frac{\text{2tp}}{\text{2tp+fp+fn}}(\tau)$ \citep{chinchor1992muc}. It is a harmonic mean of precision and recall and it was designed for unbalanced datasets. Precisely for this reason, it is sometimes~\citep{muniyandi2012network, an2015variational} used for evaluation of anomaly detection models. An issue with using the F1--score is that it puts equal weight on precision and recall. This is however in direct contrast with common practice in anomaly detection, where the focus is usually only on one of these. To compute it, we need to specify a threshold. In order to be comparable with the previously presented measures, we will compute F1--score at the threshold pertaining to a given FPR value $\alpha$
\begin{equation}
  \fa = \text{F}1(\tau) \text{ s.t. FPR}(\tau) = \alpha.   
\end{equation}

\subsection{Volume of the decision region}
\begin{figure}
\centering
\includegraphics[scale=0.8]{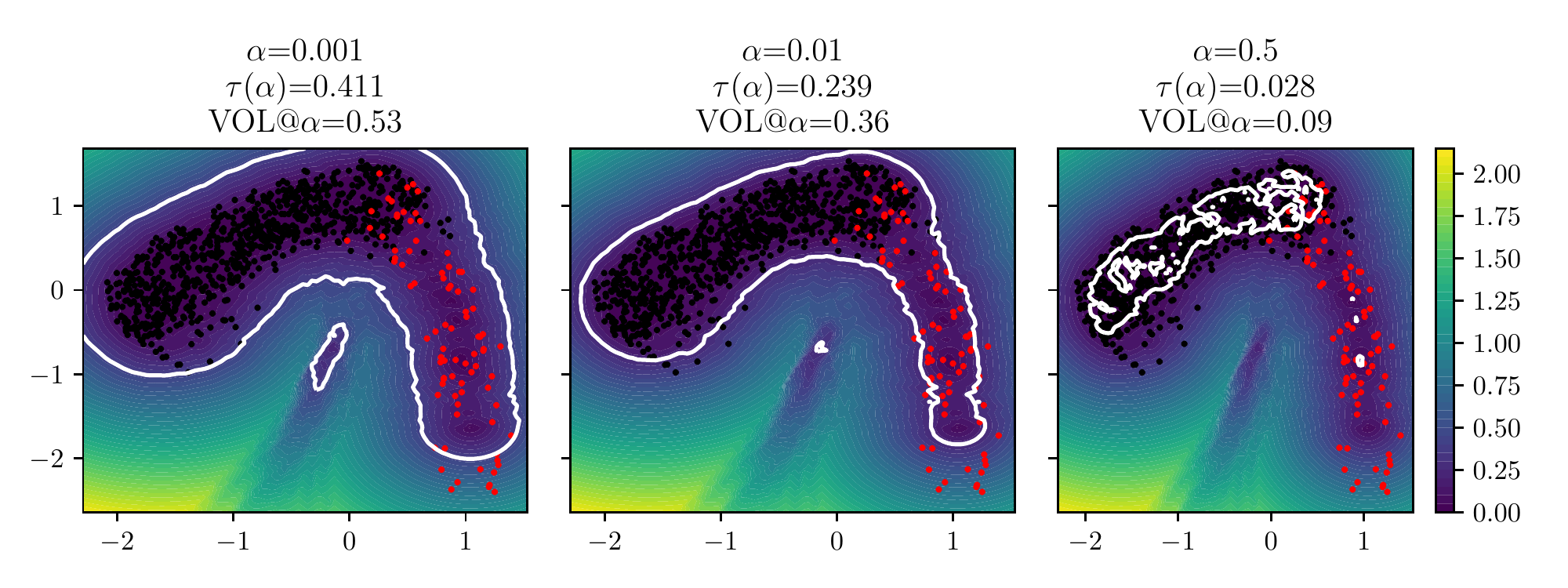}
\caption{An example of a detector and the decision region for differing values of FPR $\alpha$. The decision boundary is drawn as a white isoline at level $\tau(\alpha)$ with the estimated volume of the decision region $\text{VOL@}\alpha$, black and red dots represent normal and anomalous samples in the training set. Clearly, smaller tolerance of false positives forces us to set a higher threshold which results in a higher volume of the decision region.}
\label{fig:vol_example}
\end{figure}
All of the previous measures originate from the evaluation of the performance of binary detectors. Since semi- and unsupervised anomaly detection is closer to one-class classification or density estimation, a measure that does not require labels for its evaluation might be more useful and better describe behavior on unknown samples. If the goal is to compare two models supposed to characterize the normal class, it makes sense to choose the model enclosing the training data more tightly. This corresponds to calculating the volume inside the model's decision boundary in a similar fashion to \citep{clemenccon2013scoring}, where a theoretical justification is given. This decision boundary can be chosen to correspond to a certain level of false positive rate. We define the volume of the decision region as
\begin{equation}
  \text{VOL}(\alpha) = \int_{\mathcal{X}} \mathds{1}_{\lbrace x\in\mathcal{X}|f(x) <= \tau\rbrace} \left( x \right) dx  \text{ s.t. } \text{FPR}(\tau)=\alpha,
\end{equation}
where $\mathcal{X}$ is the input space, $f(x)$ is a decision function, $\tau$ is the decision boundary (threshold) and $\alpha$ is a given false positive rate. In other words, $\text{VOL}(\alpha)$ is the volume of subset of input space where classifier returns "normal" answer. An example of a model and its decision region for different values of $\alpha$ is shown in Fig.~\ref{fig:vol_example}. It should be noted that the idea of minimizing the volume of the decision region is native to some models, e.g. the OC-SVM algorithm \citep{scholkopf2001estimating}.

\begin{algorithm}
    \caption{Volume of the decision region computation}
    \label{alg:vol}
    \begin{algorithmic}[1]
    \Require{FPR $\alpha \in \left[0,1\right]$, a set of samples $X=\lbrace x_i \rbrace \in\mathbb{R}^d$, vector of true labels $y=\lbrace y_i\rbrace$, vector of anomaly scores $a=\lbrace a_i\rbrace$, number of samples $N \in \mathbb{N}$, a decision function $f(x)$}
    \State{$\text{TPR},\text{FPR} \gets$ Compute the ROC curve from $y$ and $a$.}
    \State{$\tau \gets$ Interpolate the ROC curve using $\alpha$ to estimate the corresponding decision threshold}
    \State$b_{min} \gets \left\{ \min_{i}(x_{ij}),j\in\hat{d}\right\}$ lower sampling boundaries
    \State$b_{max} \gets \left\{ \max_{i}(x_{ij}),j\in\hat{d}\right\}$ upper sampling boundaries
    \State{$k \gets 0$} Normal samples counter
    \For{$n\in\lbrace1,2,\dots,N\rbrace$}
      \State{$x \gets $ Sample drawn from uniform distribution $\mathcal{U}(b_{min}$,$b_{max})$}
      \State{$a_x \gets f(x)$ anomaly score of $x$}
      \If{$a_x<\tau$}
        \State{$k \gets k+1$}
      \EndIf
    \EndFor
    \State{\textbf{return} $\text{VOL@}\alpha=\frac{k}{N}$}
    \end{algorithmic}
\end{algorithm}

Computing the empirical $\text{VOL}(\alpha)$ in data space $\mathcal{X}$ is difficult and is numerically estimated by Monte-Carlo methods. In this work, we have adopted the algorithm as described in Algorithm~\ref{alg:vol}. From the definition, it is clear that the empirical measure is already normalized and can be compared across models as long as the input set of samples $X$ or the sampling boundaries are kept fixed. A preferable model yields a lower \vola. Therefore the search for an optimal model is a minimization problem. This is in contrast with the measures that were described above, which are always maximized. To make the further analysis simpler, we define the complementary normalized volume outside of the decision boundary
  \begin{equation}
    \cvola = 1-\vola
  \end{equation}

The main issue with VOL@$\alpha$ is the computational cost. Firstly, the number of samples required to cover $r$--dimensional sample space grows exponentially with $r$. Secondly, it requires $N$ evaluations of the anomaly score, which might be prohibitively expensive for some models. The first issue has been addressed in \citep{goix2016evaluate}, where the volume is computed multiple times for different subsampling of input features, however this does not seem to be optimal as it requires training a new model for each subset of features, therefore it has not been done in this paper.

\subsection{Area under the PR curve}
AUPRC, or Area Under the Precision--Recall Curve, is given by computing precision (see the definition in \ref{preca}) and recall $\text{REC}(\tau) = \frac{\text{tp}}{\text{p}}(\tau)$ for different values of classification threshold $\tau$ and then integrating the area under the resulting curve. It is however not used very often for evaluation of anomaly detection problems as it has some serious drawbacks. A PR curve has at most as many unique recall values as positive samples in the dataset. This is problematic for anomaly detection, where the number of anomalies is low, which leads to a very sparse estimate of the true PR curve. Also, our experiments make extensive comparison across different datasets with varying amount of anomalies, to which a PR curve is very sensitive. In fact, using the same trained anomaly detector and changing the contamination rate of a testing dataset produces different AUPRC results, which then makes any analysis based on AUPRC useless when the true contamination rate is unknown. Furthermore, a correct PR curve lacks a universal starting point unlike ROC, because precision is undefined for zero recall, making the computation and normalization of the area under the PR curve and the comparison between datasets even more complicated. Further critique of AUPRC compared to ROC--based measures are described in \citep{flach2015precision} -- there is no universal baseline given by a completely random detector, PR curves cannot be interpolated, the AUPRC has no other than geometric meaning etc. Finally, it has been shown in \citep{davis2006relationship} that there exists a one--to--one mapping between PR and ROC curves. Therefore we omit AUPRC from our experiments.

\section{Datasets}
\label{sec:datasets}
For the following analysis, benchmark datasets were prepared following the methodology described in \citep{emmott2013systematic}. We have used the same 36 datasets from the UCI repository \citep{Dua:2017} as in \citep{pevny2016loda}. The original datasets were either regression, binary or multiclass classification problems and were transformed into a labeled collection of normal and anomalous samples. For details on the transformation please refer to \citep{emmott2013systematic}.

The 21 multi-class classification datasets were split further --- the samples from the largest class were selected to represent the normal samples and kept fixed. The remaining classes were then added to the normal samples one-by-one and labeled as anomalous. For a $K$--class problem this procedure resulted in $K-1$ benchmark datasets with the same normal samples but different anomalous samples. Therefore, a total of 172 distinct datasets was available for the experiments with varying contamination rates (the proportion of anomalies with respect to normal samples in the training set), difficulties and topologies. For a closer description of the datasets see \citep{pevny2016loda}. The whole bulk of the datasets was not used for all experiments -- this will be made clear in the further text. 

We have also created a separate database of datasets from the original ones by reducing their dimensionality to 2 using the recently proposed UMAP transform \citep{mcinnes2018umap}. The dimensionality reduction was done because of the complexity of computing \vola\ in high dimensional spaces and also in order to enable visual analysis of the training and prediction results. In 2D, anomaly score contours and decision boundaries can be plotted easily and the researcher might gain a better intuition into the behavior of detectors.

A public repository containing the datasets is available at \texttt{https://github.com/vitskvara/UCI.jl}.

\section{Algorithms}
Four anomaly detection algorithms were chosen to test the properties of selected measures. This section contains their brief description. The chosen algorithms are amongst the most frequently used as baselines for novelty methods and are rated favorably in comparative studies such as \citep{campos2016evaluation}, \citep{goldstein2016comparative} or \citep{lazarevic2003comparative}. Note that the primary goal of this study is not to compare algorithms but to compare the measures that rate them. 

\subsection{k-Nearest Neighbours (kNN)}
The kNN algorithm \citep{ramaswamy2000efficient} is a simple but efficient model. A measure of anomalousness is the distance of a sample $x$ to its $k$-nearest neighbors. In this paper, three different ways to compute the kNN anomaly score are going to be used in accordance with \citep{harmeling2006outliers}. These are sometimes treated as different algorithms, however we are going to treat them as hyperparameters.
\begin{itemize}
  \item $\kappa(x)$: the anomaly score is the distance between $x$ and its $k$th-nearest neighbor.
  \item $\gamma(x)$: the anomaly score is the average distance of $x$ to its $k$-nearest neighbors. 
  \item $\delta(x)$: the anomaly score is the length of the mean of the vectors pointing from $x$ to its $k$-nearest neighbours.
\end{itemize}

\subsection{Local Outlier Factor (LOF)}
The LOF algorithm \citep{breunig2000lof} is based on comparing the local density of a sample $x$ with the local density of its $k$--nearest neighbours. To correctly describe the way in which the density is defined and the anomaly score is computed, let's define $k$-distance $k\text{-dist}(x)=\max_{y \in N_k(x)} d(x,y)$, where $N_k(x)$ is the set of the $k$--nearest neighbours (which can, in this context, actually contain more than k samples if some are tied at the $k$-dist$(x)$) of $x$. Also, $d(x,y)$ is the distance between $x$ and $y$ (e.g. Euclidean distance). Then we can define reachability distance $\text{rd}_k(x,y)$ as $$\text{rd}_k(x,y)=\max \lbrace k\text{-dist}(y), d(x,y) \rbrace.$$ This formula can be used to define the \textit{local reachability density} as 
\begin{equation}
  \text{LRD}_k(x)=\frac{|N_k(x)|}{\sum_{y\in N_k(x)}\text{rd}_k(x,y)}.
\end{equation}
It is in fact the inverse of average reachability--distance of $x$ and its neighbours. Finally, anomaly score of $x$ is given by comparing the $\text{LRD}_k(.)$ of $x$ and its neighbours 
\begin{equation}
  \text{LOF}_k(x)=\frac{\sum_{y\in N_k(x)} \text{LRD}_k(y)}{\text{LRD}_k(x) |N_k(x)|}.
\end{equation}

\subsection{Isolation Forest (IF)}
Isolation Forest \citep{liu2008isolation} is an algorithm based on random forests. During training, the input space is recursively and randomly cut into multidimensional boxes so that all training data points are isolated. A single cut can be expressed in terms of a tree structure with data points on the tips of the branches. An Isolation Forest instance consists of a number of such trees. It can be shown that anomalies usually require fewer cuts to be isolated -- in other words, they are more likely to be on shorter branches. Therefore, an anomaly score is computed from the average path length from the root of the tree to the datapoint. The hyperparameter to be optimized is the number of trees $N_t$.

\subsection{One Class Support Vector Machines (OC-SVM)}
OC-SVM \citep{scholkopf2001estimating} estimates the support of the training data distribution to be able to decide whether an unlabeled sample belongs to it or not. It uses a non-linear kernel function to compute the projection from the original input space to a feature space of higher dimension, in which it is possible to linearly separate the bulk of the training data by a hyperplane. The anomaly score is then the signed distance of a point from the hyperplane.  The RBF kernel will be used and the inverse of the width of the kernel $\gamma$ will be tuned. In our experiments, the hyperparameter $\nu$, which acts as an upper bound on the number of anomalies in training data, will not be tuned and will have the value 0.5 as we believe that the value of $\gamma$ is more critical to the performance.

\section{Experiments}
\subsection{Experimental settings}
\begin{table}
	\centering
	\begin{tabular}{c | c c}
		algorithm & hyperparameter & values \\
		\hline
		kNN & distance & $\lbrace \kappa(x), \gamma(x), \delta(x) \rbrace$  \\
		    & k & $\lbrace 1, 3, 5, 7, 9, 13, 21, 31, 51 \rbrace$ \\
		LOF & k & $\lbrace 10, 20, 50 \rbrace$ \\
		IF & $N_t$ & $\lbrace 50, 100, 200 \rbrace$ \\
		OCSVM & $\gamma$ &$\lbrace 0.01, 0.05, 0.1, 0.5, 1, 5, 10, 50, 100 \rbrace$ \\
	\end{tabular}
	\caption{Overview of used hyperparameter values.}
	\label{tab:hyperparams}
\end{table}

Before the exploratory investigation, we have pre-calculated performance measures from Section~\ref{sec:measures} (AUC, \aucw, \auca, \tpra, \preca, \fa, \cvola) for a set of hyper--parameter values, which are listed in Table~\ref{tab:hyperparams} for each algorithm . The performance measures were calculated on the testing set after the model was trained on the training set. The reported \preca\ measure is an average on 10 randomly subsampled testing datasets.

Each dataset has been split into training and testing data as follows. 80\% of normal data points were used for training and 20\% for testing. The contamination rate (the relative number of anomalous samples in the training data) changed according to the experimental setup. If not implicitly mentioned, all anomalies not added to the training dataset were used for testing. 

Each experiment (that means the combination of anomaly detector, hyper-parameters, and contamination rate) was repeated 10 times for randomly selected training/testing splits with fixed seed, so the training and testing folds were the same across models and hyper-parameters. The reported measures are always averages from all 10 repetitions.

\cvola\ has been estimated from 100,000 samples. A public repository containing the experimental and evaluation code is available at \\ 
\texttt{https://github.com/vitskvara/ADMetricEvaluation.jl}.

\subsection{Comparison of models}
\begin{table} 
 \center 
 \begin{tabular}[h]{c c c c c } 
  measure & kNN & LOF & IF & OCSVM  \\ 
  \hline 
  AUC & 1.56$\pm$0.72 & 2.81$\pm$1.00 & 3.55$\pm$0.79 & 2.09$\pm$0.72  \\ 
  AUC$_w$ & 1.60$\pm$0.74 & 2.69$\pm$1.05 & 3.56$\pm$0.81 & 2.15$\pm$0.73  \\ \rule{0pt}{1.5em}
  AUC@0.05 & 1.51$\pm$0.71 & 2.72$\pm$1.05 & 3.56$\pm$0.79 & 2.21$\pm$0.65  \\ 
  precision@0.05 & 1.51$\pm$0.69 & 2.64$\pm$1.00 & 3.55$\pm$0.86 & 2.30$\pm$0.72  \\ 
  TPR@0.05 & 1.68$\pm$0.73 & 2.72$\pm$0.95 & 3.51$\pm$0.84 & 2.09$\pm$0.64  \\ 
  F1@0.05 & 1.80$\pm$0.74 & 2.78$\pm$0.90 & 3.55$\pm$0.78 & 1.87$\pm$0.69  \\ 
  CVOL@0.05 & 1.65$\pm$0.43 & 3.19$\pm$0.60 & 3.64$\pm$0.60 & 1.53$\pm$0.43  \\ \rule{0pt}{1.5em}
  AUC@0.01 & 1.58$\pm$0.71 & 2.63$\pm$1.09 & 3.59$\pm$0.74 & 2.20$\pm$0.73  \\ 
  precision@0.01 & 2.22$\pm$1.25 & 2.32$\pm$0.84 & 2.94$\pm$1.36 & 2.51$\pm$0.68  \\ 
  TPR@0.01 & 1.52$\pm$0.59 & 2.72$\pm$0.98 & 3.64$\pm$0.69 & 2.12$\pm$0.78  \\ 
  F1@0.01 & 1.76$\pm$0.70 & 2.86$\pm$0.80 & 3.70$\pm$0.59 & 1.68$\pm$0.77  \\ 
  CVOL@0.01 & 1.85$\pm$0.68 & 3.20$\pm$0.61 & 3.59$\pm$0.66 & 1.36$\pm$0.47  \\ 
 \end{tabular}
 \caption{Means and standard deviations of algorithm ranks using different measures, 0\% training contamination.} 
 \label{tab:model_ranks_0} 
\end{table}
\begin{table} 
 \center 
 \begin{tabular}[h]{c c c c c } 
  measure & kNN & LOF & IF & OCSVM  \\ 
  \hline 
  AUC & 1.53$\pm$0.64 & 2.96$\pm$0.77 & 3.61$\pm$0.72 & 1.90$\pm$0.57  \\ 
  AUC$_w$ & 1.49$\pm$0.61 & 3.02$\pm$0.64 & 3.68$\pm$0.68 & 1.81$\pm$0.56  \\ \rule{0pt}{1.5em}
  AUC@0.05 & 1.55$\pm$0.64 & 2.87$\pm$0.70 & 3.72$\pm$0.62 & 1.86$\pm$0.59  \\ 
  precision@0.05 & 1.49$\pm$0.74 & 2.93$\pm$0.67 & 3.61$\pm$0.72 & 1.96$\pm$0.50  \\ 
  TPR@0.05 & 1.60$\pm$0.69 & 2.85$\pm$0.67 & 3.44$\pm$0.73 & 2.11$\pm$0.55  \\ 
  F1@0.05 & 1.61$\pm$0.68 & 2.84$\pm$0.68 & 3.45$\pm$0.72 & 2.10$\pm$0.56  \\ 
  CVOL@0.05 & 1.21$\pm$0.41 & 2.95$\pm$0.60 & 3.84$\pm$0.39 & 2.01$\pm$0.65  \\ \rule{0pt}{1.5em}
  AUC@0.01 & 1.54$\pm$0.57 & 2.90$\pm$0.68 & 3.72$\pm$0.63 & 1.85$\pm$0.59  \\ 
  precision@0.01 & 2.11$\pm$1.16 & 2.64$\pm$0.72 & 3.08$\pm$1.25 & 2.17$\pm$0.69  \\ 
  TPR@0.01 & 1.57$\pm$0.58 & 2.85$\pm$0.71 & 3.71$\pm$0.60 & 1.87$\pm$0.60  \\ 
  F1@0.01 & 1.57$\pm$0.58 & 2.82$\pm$0.70 & 3.72$\pm$0.60 & 1.89$\pm$0.60  \\ 
  CVOL@0.01 & 1.38$\pm$0.64 & 2.96$\pm$0.58 & 3.81$\pm$0.45 & 1.85$\pm$0.68  \\ 
 \end{tabular}
 \caption{Means and standard deviations of algorithm ranks using different measures, UMAP datasets, 0\% training contamination.} 
 \label{tab:model_ranks_umap_0} 
\end{table}
Comparing different models using the approach suggested in \citep{demvsar2006statistical} amounts to ranking detectors using the measure of choice on each dataset separately and then calculating the average rank of each detector across the datasets. The best detector would have the lowest rank and \citep{demvsar2006statistical} further proposes to use a Friedman test to asses whether the difference is statistically significant.

Table~\ref{tab:model_ranks_0} shows these average ranks for all four detectors used in this study on full-dimensional datasets using all the measures. We observe that all measures except \cvola\ are correlated in the sense that all detectors have the same rank most of the time.\footnote{By the same rank we mean the rank determined from average rank.} kNN detector is always the best, followed by OC-SVM, LOF, and Isolation Forest. Measures differ mostly to the extent to which the average rank is different. The best example is precision@0.01, where all detectors have very similar ranks and the difference in their performance is not statistically significant. Should we compare the detectors using \aucw, kNN would be a clear winner by a large extent. As will be shown below, we attribute this behavior of precision@0.01 to noisy estimates of this measure.

Table~\ref{tab:model_ranks_umap_0} shows the ranks of the same anomaly detectors for the same performance measures as in Table~\ref{tab:model_ranks_0}, but estimated on a down-scaled version of datasets as described in Section~\ref{sec:datasets}. We observe that the average ranks of the compared detectors are similar to those calculated on the datasets with full dimension. This suggests that the datasets downscaled by UMAP might be good for studying the behavior of anomaly detection methods.

\cvola\ is the only measure that relies purely on samples of normal data when the contamination is not taken into the account. It is therefore not surprising that the average ranks of detectors using this measure are different. When this measure would be used, OC-SVM would be the best detector closely followed by kNN. LOF and Isolation Forests would be left far behind. Again, this should not be surprising since OC-SVM has been designed to optimize precisely this measure, yet it is difficult to select its hyper-parameters.

\begin{table*} 
 \center 
 \begin{tabular}[h]{c c c c c c c c c c c c c c c } 
  measure & \rotatebox{90}{AUC} & \rotatebox{90}{AUC$_w$} & \rotatebox{90}{AUC@0.05} & \rotatebox{90}{AUC@0.01} & \rotatebox{90}{precision@0.05} & \rotatebox{90}{precision@0.01} & \rotatebox{90}{TPR@0.05} & \rotatebox{90}{TPR@0.01} & \rotatebox{90}{F1@0.05} & \rotatebox{90}{F1@0.01} & \rotatebox{90}{CVOL@0.05} & \rotatebox{90}{CVOL@0.01} & \rotatebox{90}{} & \rotatebox{90}{mean}  \\ 
  \hline 
  AUC & -- & \cellcolor{gray!45}0.84 & 0.77 & 0.67 & \cellcolor{gray!15}0.75 & 0.63 & 0.69 & 0.65 & 0.62 & 0.59 & 0.42 & 0.46 &  & 0.67  \\ 
  AUC$_w$ & \cellcolor{gray!45}0.84 & -- & \cellcolor{gray!30}0.82 & 0.75 & \cellcolor{gray!30}0.76 & \cellcolor{gray!15}0.70 & 0.68 & 0.70 & 0.62 & 0.63 & 0.42 & 0.49 &  & 0.70  \\ 
  AUC@0.05 & \cellcolor{gray!30}0.77 & \cellcolor{gray!30}0.82 & -- & \cellcolor{gray!30}0.83 & \cellcolor{gray!45}0.77 & \cellcolor{gray!45}0.72 & \cellcolor{gray!30}0.78 & \cellcolor{gray!15}0.82 & \cellcolor{gray!30}0.72 & \cellcolor{gray!15}0.73 & 0.47 & 0.55 &  & \cellcolor{gray!45}0.75  \\ 
  AUC@0.01 & 0.67 & 0.75 & \cellcolor{gray!45}0.83 & -- & 0.68 & \cellcolor{gray!30}0.70 & 0.64 & \cellcolor{gray!45}0.89 & 0.61 & \cellcolor{gray!30}0.78 & 0.43 & 0.57 &  & \cellcolor{gray!15}0.71  \\ 
  precision@0.05 & \cellcolor{gray!15}0.75 & \cellcolor{gray!15}0.76 & 0.77 & 0.68 & -- & 0.65 & \cellcolor{gray!15}0.70 & 0.68 & 0.65 & 0.63 & 0.44 & 0.49 &  & 0.68  \\ 
  precision@0.01 & 0.63 & 0.70 & 0.72 & 0.70 & 0.65 & -- & 0.56 & 0.67 & 0.47 & 0.57 & 0.32 & 0.44 &  & 0.62  \\ 
  TPR@0.05 & 0.69 & 0.68 & 0.78 & 0.64 & 0.70 & 0.56 & -- & 0.67 & \cellcolor{gray!45}0.91 & 0.61 & 0.48 & 0.47 &  & 0.68  \\ 
  TPR@0.01 & 0.65 & 0.70 & \cellcolor{gray!15}0.82 & \cellcolor{gray!45}0.89 & 0.68 & 0.67 & 0.67 & -- & 0.63 & \cellcolor{gray!45}0.88 & 0.45 & \cellcolor{gray!15}0.60 &  & \cellcolor{gray!30}0.72  \\ 
  F1@0.05 & 0.62 & 0.62 & 0.72 & 0.61 & 0.65 & 0.47 & \cellcolor{gray!45}0.91 & 0.63 & -- & 0.67 & \cellcolor{gray!30}0.54 & 0.52 &  & 0.66  \\ 
  F1@0.01 & 0.59 & 0.63 & 0.73 & \cellcolor{gray!15}0.78 & 0.63 & 0.57 & 0.61 & \cellcolor{gray!30}0.88 & \cellcolor{gray!15}0.67 & -- & \cellcolor{gray!15}0.48 & \cellcolor{gray!45}0.70 &  & 0.69  \\ 
  CVOL@0.05 & 0.42 & 0.42 & 0.47 & 0.43 & 0.44 & 0.32 & 0.48 & 0.45 & 0.54 & 0.48 & -- & \cellcolor{gray!30}0.64 &  & 0.51  \\ 
  CVOL@0.01 & 0.46 & 0.49 & 0.55 & 0.57 & 0.49 & 0.44 & 0.47 & 0.60 & 0.52 & 0.70 & \cellcolor{gray!45}0.64 & -- &  & 0.58  \\ 
 \end{tabular}
 \caption{Average of Kendall correlation between measures over datasets, 0\% contamination. Level of shading highlights three highest correlations in a column.} 
 \label{tab:measure_correlation_full_0} 
\end{table*}
Finally, we have compared the similarity of measures using the Kendall rank correlation coefficient $R$ \citep{kendall1938new}. We have computed $R_{ij}=R(y_i,y_j)$, where $y_i$ is the vector of values of the $i$-th measure computed on a single dataset for different models and hyperparameters, and $i,j$ run over all possible measure combinations. The correlation coefficients averaged across all datasets (see Table~\ref{tab:measure_correlation_full_0}) reveal disturbing phenomena.  While \aucw\ and \auca\ are relatively well correlated with all other measures, which supports the similar ranks in Table~\ref{tab:model_ranks_0}, there is a surprisingly weak correlation of \preca\ and \tpra, even in the case of the same measures differing only in thresholds, which is little alarming, as these are the measures that the practitioners are interested in.

\paragraph{Conclusion:}
From the perspective of comparison of detectors, AUC does not seem to be a particularly bad choice, as average ranks used in comparison of detectors according to \citep{demvsar2006statistical} provide similar results. This would suggest that cases demonstrated in Fig.~\ref{fig:bad_auc} rarely happen in reality, although we cannot completely rule them out.

\subsection{Selection of models}
\begin{table*} 
 \center 
 \resizebox{\textwidth}{!}{ 
 \begin{tabular}[h]{c c c c c c c c c c c c c c c } 
  max/loss & \rotatebox{90}{AUC} & \rotatebox{90}{AUC$_w$} & \rotatebox{90}{AUC@0.05} & \rotatebox{90}{AUC@0.01} & \rotatebox{90}{precision@0.05} & \rotatebox{90}{precision@0.01} & \rotatebox{90}{TPR@0.05} & \rotatebox{90}{TPR@0.01} & \rotatebox{90}{F1@0.05} & \rotatebox{90}{F1@0.01} & \rotatebox{90}{CVOL@0.05} & \rotatebox{90}{CVOL@0.01} & \rotatebox{90}{} & \rotatebox{90}{mean}  \\ 
  \hline 
  AUC & -- & \cellcolor{gray!30}0.4\% & 1.2\% & 2.0\% & 2.0\% & 2.3\% & \cellcolor{gray!15}1.0\% & 2.4\% & 7.1\% & 10.9\% & 3.7\% & 8.9\% &  & 3.5\%  \\ 
  AUC$_w$ & \cellcolor{gray!45}0.1\% & -- & \cellcolor{gray!45}0.5\% & 1.3\% & \cellcolor{gray!15}1.7\% & \cellcolor{gray!30}1.6\% & \cellcolor{gray!30}0.6\% & 1.6\% & 6.7\% & 10.3\% & 3.6\% & 9.7\% &  & 3.1\%  \\ 
  AUC@0.05 & \cellcolor{gray!30}0.3\% & \cellcolor{gray!45}0.2\% & -- & \cellcolor{gray!30}0.7\% & \cellcolor{gray!45}1.5\% & 1.8\% & \cellcolor{gray!45}0.4\% & \cellcolor{gray!30}0.9\% & \cellcolor{gray!30}5.3\% & 8.4\% & 1.9\% & 8.0\% &  & \cellcolor{gray!45}2.4\%  \\ 
  AUC@0.01 & 0.7\% & 0.8\% & \cellcolor{gray!15}1.0\% & -- & 2.5\% & \cellcolor{gray!45}1.4\% & 2.1\% & \cellcolor{gray!45}0.5\% & 6.3\% & \cellcolor{gray!30}7.9\% & 1.7\% & \cellcolor{gray!15}5.1\% &  & \cellcolor{gray!30}2.5\%  \\ 
  precision@0.05 & 0.5\% & \cellcolor{gray!15}0.7\% & 1.0\% & 2.3\% & -- & 2.4\% & 1.3\% & 2.2\% & 6.6\% & 10.1\% & \cellcolor{gray!15}1.5\% & 7.1\% &  & 3.0\%  \\ 
  precision@0.01 & 0.7\% & 1.0\% & 1.5\% & \cellcolor{gray!15}0.7\% & \cellcolor{gray!30}1.6\% & -- & 2.8\% & \cellcolor{gray!15}1.4\% & 10.1\% & 12.5\% & \cellcolor{gray!45}0.0\% & \cellcolor{gray!45}1.9\% &  & 2.9\%  \\ 
  TPR@0.05 & \cellcolor{gray!15}0.3\% & 1.5\% & 1.9\% & 5.7\% & 2.2\% & 4.6\% & -- & 4.5\% & 6.2\% & 12.3\% & 3.1\% & 9.2\% &  & 4.3\%  \\ 
  TPR@0.01 & 0.6\% & 1.1\% & \cellcolor{gray!30}1.0\% & \cellcolor{gray!45}0.5\% & 2.3\% & \cellcolor{gray!15}1.8\% & 2.2\% & -- & \cellcolor{gray!15}6.2\% & \cellcolor{gray!15}8.2\% & 1.7\% & 8.4\% &  & \cellcolor{gray!15}2.8\%  \\ 
  F1@0.05 & 3.6\% & 11.0\% & 10.3\% & 13.1\% & 14.0\% & 21.2\% & 8.8\% & 12.4\% & -- & \cellcolor{gray!45}6.0\% & 1.7\% & 12.1\% &  & 9.5\%  \\ 
  F1@0.01 & 3.5\% & 17.7\% & 16.9\% & 27.7\% & 13.8\% & 37.1\% & 10.1\% & 24.5\% & \cellcolor{gray!45}3.4\% & -- & 1.5\% & \cellcolor{gray!30}3.1\% &  & 13.3\%  \\ 
  CVOL@0.05 & 3.8\% & 13.3\% & 18.1\% & 22.1\% & 18.7\% & 15.8\% & 15.9\% & 19.7\% & 9.5\% & 22.0\% & -- & 14.4\% &  & 14.4\%  \\ 
  CVOL@0.01 & 1.4\% & 6.4\% & 7.0\% & 12.0\% & 6.5\% & 9.3\% & 5.9\% & 7.7\% & 7.9\% & 9.2\% & \cellcolor{gray!30}0.2\% & -- &  & 6.1\%  \\ 
 \end{tabular}
 }
 \caption{Means of relative loss in a column measure when optimal model and hyperparameters are selected using the row measure. 0\%  training contamination. Level of shading highlights three best results in a column.} 
 \label{tab:measure_comparison_full_0_by_models} 
\end{table*}
\begin{table*} 
 \center 
 \resizebox{\textwidth}{!}{ 
 \begin{tabular}[h]{c c c c c c c c c c c c c c c } 
  max/loss & \rotatebox{90}{AUC} & \rotatebox{90}{AUC$_w$} & \rotatebox{90}{AUC@0.05} & \rotatebox{90}{AUC@0.01} & \rotatebox{90}{precision@0.05} & \rotatebox{90}{precision@0.01} & \rotatebox{90}{TPR@0.05} & \rotatebox{90}{TPR@0.01} & \rotatebox{90}{F1@0.05} & \rotatebox{90}{F1@0.01} & \rotatebox{90}{CVOL@0.05} & \rotatebox{90}{CVOL@0.01} & \rotatebox{90}{} & \rotatebox{90}{mean}  \\ 
  \hline 
  AUC & -- & 1.2\% & 2.1\% & 5.1\% & 2.5\% & 3.5\% & \cellcolor{gray!15}1.3\% & 4.3\% & 8.1\% & 13.1\% & 4.1\% & 10.1\% &  & 4.6\%  \\ 
  AUC$_w$ & \cellcolor{gray!45}0.2\% & -- & \cellcolor{gray!45}0.6\% & 1.2\% & \cellcolor{gray!30}1.9\% & \cellcolor{gray!15}1.2\% & \cellcolor{gray!30}1.1\% & 1.6\% & 7.6\% & 11.1\% & 4.1\% & 10.4\% &  & 3.4\%  \\ 
  AUC@0.05 & 0.7\% & \cellcolor{gray!45}0.4\% & -- & \cellcolor{gray!30}0.7\% & \cellcolor{gray!45}1.8\% & \cellcolor{gray!30}1.1\% & \cellcolor{gray!45}0.9\% & \cellcolor{gray!30}0.9\% & \cellcolor{gray!30}6.2\% & 9.5\% & 2.4\% & 9.7\% &  & \cellcolor{gray!30}2.9\%  \\ 
  AUC@0.01 & 0.9\% & \cellcolor{gray!30}0.8\% & \cellcolor{gray!30}0.8\% & -- & 2.3\% & \cellcolor{gray!45}1.0\% & 2.2\% & \cellcolor{gray!45}0.5\% & 7.1\% & \cellcolor{gray!30}8.7\% & 2.2\% & \cellcolor{gray!15}6.3\% &  & \cellcolor{gray!45}2.7\%  \\ 
  precision@0.05 & \cellcolor{gray!15}0.5\% & \cellcolor{gray!15}1.0\% & 1.5\% & 3.2\% & -- & 2.3\% & 1.4\% & 2.7\% & 7.5\% & 11.7\% & 2.1\% & 7.5\% &  & 3.5\%  \\ 
  precision@0.01 & 1.2\% & 1.3\% & 1.7\% & \cellcolor{gray!15}1.2\% & \cellcolor{gray!15}2.1\% & -- & 3.1\% & \cellcolor{gray!15}1.4\% & 11.2\% & 14.2\% & \cellcolor{gray!45}0.1\% & \cellcolor{gray!45}3.9\% &  & 3.4\%  \\ 
  TPR@0.05 & \cellcolor{gray!30}0.4\% & 2.2\% & 2.5\% & 7.9\% & 2.4\% & 5.2\% & -- & 6.1\% & \cellcolor{gray!15}6.7\% & 13.6\% & 3.5\% & 8.9\% &  & 4.9\%  \\ 
  TPR@0.01 & 0.9\% & 1.2\% & \cellcolor{gray!15}0.9\% & \cellcolor{gray!45}0.6\% & 2.6\% & 1.3\% & 2.5\% & -- & 7.1\% & \cellcolor{gray!15}9.0\% & 2.1\% & 9.6\% &  & \cellcolor{gray!15}3.2\%  \\ 
  F1@0.05 & 4.5\% & 12.6\% & 10.8\% & 13.7\% & 16.5\% & 19.3\% & 10.2\% & 12.6\% & -- & \cellcolor{gray!45}6.2\% & 2.0\% & 11.2\% &  & 10.0\%  \\ 
  F1@0.01 & 7.2\% & 21.9\% & 21.3\% & 19.3\% & 27.2\% & 33.6\% & 22.7\% & 20.7\% & \cellcolor{gray!45}4.3\% & -- & \cellcolor{gray!15}1.9\% & \cellcolor{gray!30}4.3\% &  & 15.4\%  \\ 
  CVOL@0.05 & 4.6\% & 17.5\% & 24.2\% & 31.3\% & 20.5\% & 23.8\% & 20.0\% & 27.6\% & 12.6\% & 27.3\% & -- & 14.1\% &  & 18.6\%  \\ 
  CVOL@0.01 & 3.3\% & 14.8\% & 19.0\% & 26.3\% & 14.0\% & 23.7\% & 15.1\% & 22.6\% & 13.7\% & 18.7\% & \cellcolor{gray!30}0.2\% & -- &  & 14.3\%  \\ 
 \end{tabular}
 }
 \caption{Means of relative loss in a column measure when optimal model and hyperparameters are selected using the row measure. 1\% contamination.} 
 \label{tab:measure_comparison_full_1_by_models} 
\end{table*}
\begin{table*} 
 \center 
 \resizebox{\textwidth}{!}{ 
 \begin{tabular}[h]{c c c c c c c c c c c c c c c } 
  max/loss & \rotatebox{90}{AUC} & \rotatebox{90}{AUC$_w$} & \rotatebox{90}{AUC@0.05} & \rotatebox{90}{AUC@0.01} & \rotatebox{90}{precision@0.05} & \rotatebox{90}{precision@0.01} & \rotatebox{90}{TPR@0.05} & \rotatebox{90}{TPR@0.01} & \rotatebox{90}{F1@0.05} & \rotatebox{90}{F1@0.01} & \rotatebox{90}{CVOL@0.05} & \rotatebox{90}{CVOL@0.01} & \rotatebox{90}{} & \rotatebox{90}{mean}  \\ 
  \hline 
  AUC & -- & \cellcolor{gray!15}1.3\% & 3.2\% & 5.4\% & 2.7\% & 5.1\% & 1.9\% & 5.2\% & 10.8\% & 21.2\% & 4.0\% & 9.7\% &  & 5.9\%  \\ 
  AUC$_w$ & \cellcolor{gray!45}0.3\% & -- & \cellcolor{gray!45}0.8\% & 1.7\% & \cellcolor{gray!45}1.8\% & \cellcolor{gray!45}1.5\% & \cellcolor{gray!30}1.0\% & 1.9\% & 9.7\% & 18.3\% & 3.3\% & 8.3\% &  & \cellcolor{gray!30}4.1\%  \\ 
  AUC@0.05 & 0.8\% & \cellcolor{gray!45}0.6\% & -- & \cellcolor{gray!30}1.6\% & \cellcolor{gray!30}1.9\% & \cellcolor{gray!30}1.8\% & \cellcolor{gray!45}0.9\% & \cellcolor{gray!30}1.1\% & \cellcolor{gray!15}8.4\% & 17.2\% & 2.2\% & 9.0\% &  & \cellcolor{gray!45}3.8\%  \\ 
  AUC@0.01 & 1.9\% & 2.2\% & \cellcolor{gray!15}1.6\% & -- & 3.8\% & \cellcolor{gray!15}2.0\% & 3.7\% & \cellcolor{gray!45}0.3\% & 10.5\% & \cellcolor{gray!30}16.1\% & 1.9\% & \cellcolor{gray!15}6.5\% &  & \cellcolor{gray!15}4.2\%  \\ 
  precision@0.05 & \cellcolor{gray!15}0.6\% & \cellcolor{gray!30}0.9\% & 1.6\% & 3.3\% & -- & 2.4\% & \cellcolor{gray!15}1.3\% & 3.0\% & 9.6\% & 19.4\% & 2.0\% & 6.9\% &  & 4.3\%  \\ 
  precision@0.01 & 2.3\% & 2.6\% & 1.8\% & \cellcolor{gray!15}1.6\% & 2.9\% & -- & 4.2\% & \cellcolor{gray!15}1.5\% & 15.3\% & 24.5\% & \cellcolor{gray!45}0.1\% & \cellcolor{gray!45}2.8\% &  & 5.0\%  \\ 
  TPR@0.05 & \cellcolor{gray!30}0.6\% & 1.5\% & 2.0\% & 5.6\% & \cellcolor{gray!15}2.1\% & 4.6\% & -- & 4.6\% & \cellcolor{gray!30}8.2\% & 19.3\% & 2.7\% & 7.7\% &  & 4.9\%  \\ 
  TPR@0.01 & 1.6\% & 2.1\% & \cellcolor{gray!30}1.5\% & \cellcolor{gray!45}0.5\% & 3.9\% & 2.3\% & 3.7\% & -- & 10.4\% & \cellcolor{gray!15}16.2\% & \cellcolor{gray!15}1.9\% & 8.0\% &  & 4.3\%  \\ 
  F1@0.05 & 7.8\% & 19.4\% & 16.2\% & 16.0\% & 26.2\% & 28.9\% & 17.5\% & 16.0\% & -- & \cellcolor{gray!45}6.9\% & 1.9\% & 10.6\% &  & 14.0\%  \\ 
  F1@0.01 & 12.6\% & 32.9\% & 29.9\% & 24.6\% & 39.3\% & 50.8\% & 32.9\% & 27.4\% & \cellcolor{gray!45}5.1\% & -- & 1.9\% & \cellcolor{gray!30}4.8\% &  & 21.9\%  \\ 
  CVOL@0.05 & 8.5\% & 26.4\% & 36.5\% & 41.1\% & 27.9\% & 38.5\% & 32.7\% & 39.2\% & 23.9\% & 41.4\% & -- & 12.5\% &  & 27.4\%  \\ 
  CVOL@0.01 & 8.0\% & 25.1\% & 34.0\% & 37.4\% & 23.8\% & 43.9\% & 30.6\% & 36.8\% & 26.8\% & 36.8\% & \cellcolor{gray!30}0.3\% & -- &  & 25.3\%  \\ 
 \end{tabular}
 }
 \caption{Means of relative performance loss in a column measure when optimal model and hyperparameters are selected using the row measure. 5\% training contamination.} 
 \label{tab:measure_comparison_full_5_by_models} 
\end{table*}

Let us now investigate the performance measures from the point of view of a user who wishes to select a detector for a particular application scenario. The user does a review of the prior art, noticing performance measures in papers (typically AUC), and has to decide which detector(s) to implement and use. As explained in the introduction, the mismatch between the selection measure (AUC) and the measure well describing the application conditions (\tpra\ or \preca) is almost certain.

This problem is captured in Table~\ref{tab:measure_comparison_full_0_by_models} showing the average relative performance loss calculated as $\frac{c_{\mathrm{best}} - c_{\mathrm{used}}}{c_{\mathrm{best}}},$ where $c_{\mathrm{best}}$ is the best possible value of a measure important for the particular problem and $c_{\mathrm{used}}$ is the value of the same measure for a particular detector chosen with another performance measure. Therefore lower is better. In Table~\ref{tab:measure_comparison_full_0_by_models}, the rows correspond to the measure used to select the detector and the columns correspond to measures important for the application. The anomaly detection algorithm is considered to be a hyper-parameter in Table~\ref{tab:measure_comparison_full_0_by_models}, which means that the user chooses the best algorithms and its hyper-parameters on the testing set. The reason for choosing hyper-parameters on the testing set is that we want to isolate all sources of noise and focus on mismatch between the tested measures.

The table clearly shows that in case of a mismatch between the reported and application measures, \aucw\ and AUC@0.05 result in consistently smaller relative errors (better performance in the application) than the AUC. Taking \fa\ and \cvola\ aside, whose importance for real application is dubious and the estimation can be noisy, AUC seems like the worst choice of measure, as all the others are better. This inferiority of AUC is even more pronounced in the case when the training data are not clean but contaminated by anomalous samples, as is shown in Table~\ref{tab:measure_comparison_full_1_by_models} and~\ref{tab:measure_comparison_full_5_by_models}.

\begin{figure}
\centering
  \includegraphics[scale=0.85]{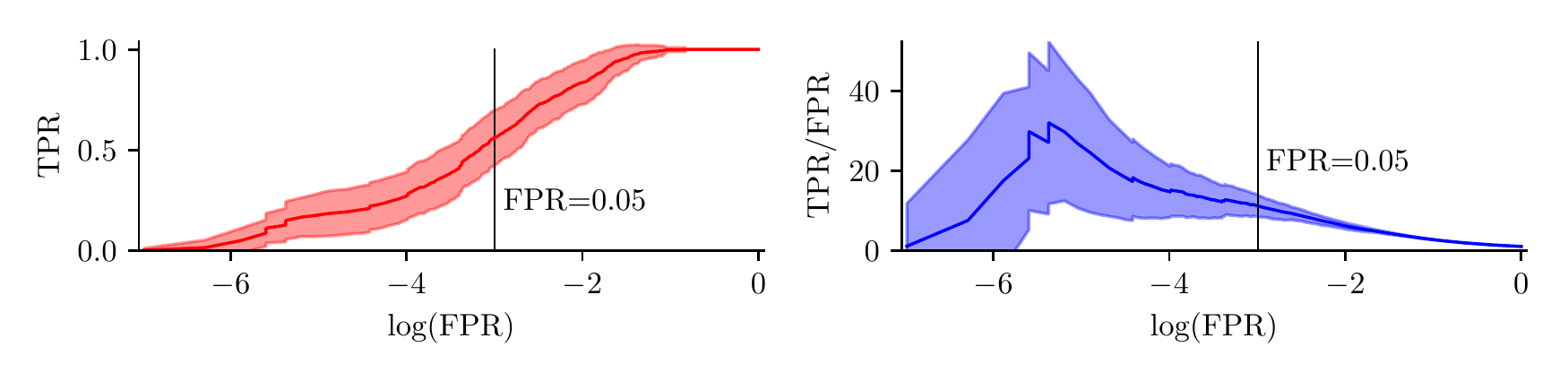}
  \caption{Mean of 100 ROC curves with one standard deviation interval in red, mean of TPR/FPR with one standard deviation in blue. Since \aucw\ integrates over FPR$^{-1}$, the noise in data has a more pronounced effect on its estimates in areas of low FPR values. Data obtained on the abalone dataset using kNN model with $\delta$ distance and $k=31$.}
  \label{fig:ROC_mc_experiment_horizontal}
\end{figure}

The result that AUC is less informative of the quality of anomaly detectors performing on low false positive rates is expectable. What is rather surprising is that AUC@0.05 seems to be consistently a good choice, even if we are interested in different false positive rates, e.g. at 0.01. We have expected \aucw\ to be more versatile. We believe that the superiority of \auca\ is caused by (i) being a crude estimate of \aucw\ and (ii) being robust. Specifically, estimates of a fraction of true and false positive rate in the calculation of \aucw~\eqref{eq:AUCw} are very noisy for low false positive rates, as is shown in Fig.~\ref{fig:ROC_mc_experiment_horizontal}.

\paragraph{Conclusion:}
Choosing the detector on the basis of AUC does not seems to be the best, as using \auca\ and \aucw\ seems to consistently yield in better performance in all application measures. \aucw\ is theoretically justified by its connection to Neyman-Pearson classifier \citep{scott2007performance}, but experiments suggest AUC@0.05 to be slightly better. We believe this to be due to the robustness of AUC@0.05 and/or instability of \aucw\ which was demonstrated in Fig.~\ref{fig:ROC_mc_experiment_horizontal}.

\subsection{Influence of unrepresentative examples of anomalies}

\begin{figure}
\centering
\includegraphics[width=0.6\columnwidth]{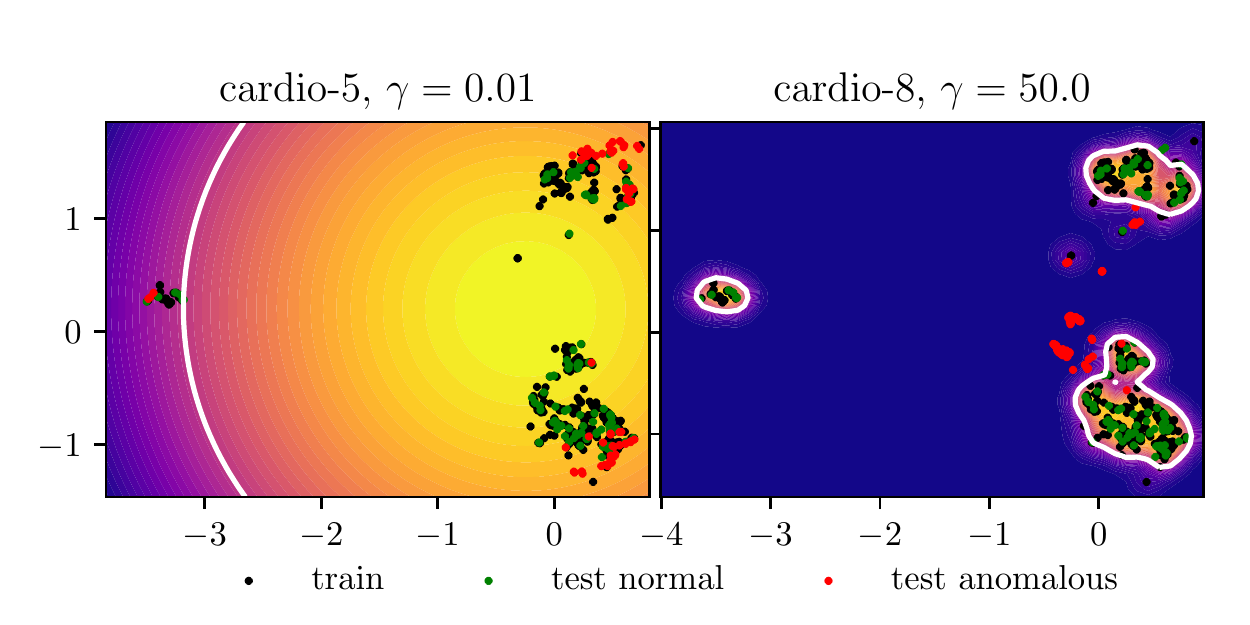}
\caption{An example of fitting the same model (OC-SVM) on the same normal data (black dots) for two different settings of hyper-parameter (width of Gaussian kernel $\gamma$) and evaluated on two different anomaly classes (red dots). Green dots show testing normal samples. Decision boundary is shown in white and background is proportionate to anomaly score.}
\label{fig:cardio_example}

\vspace{1cm}
\begin{tabular}{l  c c c c c}
   & \multicolumn{2}{c}{cardio-5} & &\multicolumn{2}{c}{cardio-8} \\
  \cline{2-3} \cline{5-6}
  parameter & AUC & CVOL@0.05 & &AUC & CVOL@0.05 \\
  \hline
  $\gamma = 0.01$ & 0.87 & 0.10&  & 0.02 & 0.11  \\
  $\gamma = 50.0$ & 0.74 & 0.91&  & 0.99 & 0.91 \\
\end{tabular}
\captionof{table}{Measures of OC-SVM fitted on Cardiotocography class 2 and evaluated on classes 5 and 8.}
\label{tab:cardio_example}
\end{figure}

As already mentioned above, most publications on anomaly detectors evaluate and compare them using supervised measures. Albeit not stated explicitly, this implies that the user possesses a set of anomalies with similar probability distribution he is interested in or he expects to occur. When validation samples are not representative, the consequences can be serious. This is especially likely in non-stationary environments like computer security, where the field is constantly evolving and anomalies known today are not representative examples of anomalies in the future. An example of this phenomenon is shown in Fig.~\ref{fig:cardio_example}, where two OC-SVMs differing in the values of a hyper-parameter (width of Gaussian kernel) have been fitted on a class 2 and evaluated on classes 5 and 8. According to AUCs shown in Table~\ref{tab:cardio_example} one value of hyper-parameter is better for anomalies from class 5, but it leads to terrible results on anomalies from class 8 and vice versa. 

This problem can be mitigated by replacing measures requiring anomalous samples (AUC, precision@$\alpha$, TPR@$\alpha$) by measures relying only on the normal data, for example the volume of the decision region used here and proposed in \citep{steinwart2005classification}. Indeed, in the above example, should we have selected the detector purely based on the volume, we would prevent the catastrophic result (as we would choose the detector with $\gamma = 50.0$). Of course in the case when anomalies available for evaluation of the detector are representative of anomalies we wish to detect/we encounter in real, we would likely lose some performance, as we would play a safe strategy.

To study the impact of different distributions of anomalies in the validation and testing phase, and robustness of performance measures, we conducted the following experiment. For each multi-class dataset in our repository, we have trained anomaly detectors on a single fixed class (the one with the highest number of samples), and evaluated detectors using anomalies from the remaining classes separately. Then, we would select the best detector and its hyper-parameters using a particular measure and class of anomalies, and evaluate this detector on the remaining classes of anomalies (and all measures). Again, we report the average performance loss of this anomaly-class mismatch.

\begin{table} 
 \center 
 \begin{tabular}[h]{c c c c c c c c } 
  measure & \rotatebox{90}{AUC} & \rotatebox{90}{AUC$_w$} & \rotatebox{90}{AUC@0.05} & \rotatebox{90}{precision@0.05} & \rotatebox{90}{TPR@0.05} & \rotatebox{90}{F1@0.05} & \rotatebox{90}{CVOL@0.05}  \\ 
  \hline 
  AUC & 2.0\% & 3.9\% & 6.5\% & 5.4\% & 5.3\% & 4.6\% & 5.2\%  \\ 
  AUC$_w$ & \cellcolor{gray!30}1.6\% & \cellcolor{gray!30}2.8\% & \cellcolor{gray!30}4.7\% & \cellcolor{gray!30}4.2\% & \cellcolor{gray!30}4.2\% & \cellcolor{gray!30}3.4\% & \cellcolor{gray!15}4.4\%  \\ 
  AUC@0.05 & 1.7\% & \cellcolor{gray!15}3.1\% & \cellcolor{gray!15}4.8\% & \cellcolor{gray!15}4.6\% & 4.5\% & 3.7\% & \cellcolor{gray!30}4.3\%  \\ 
  precision@0.05 & 1.8\% & 3.4\% & 5.7\% & 4.8\% & 4.8\% & 3.9\% & 4.7\%  \\ 
  TPR@0.05 & \cellcolor{gray!15}1.7\% & 3.8\% & 6.5\% & 5.0\% & \cellcolor{gray!15}4.2\% & 3.5\% & 4.9\%  \\ 
  F1@0.05 & 1.7\% & 3.8\% & 6.4\% & 5.0\% & 4.2\% & \cellcolor{gray!15}3.5\% & 4.9\%  \\ 
  CVOL@0.05 & \cellcolor{gray!45}1.4\% & \cellcolor{gray!45}2.4\% & \cellcolor{gray!45}3.8\% & \cellcolor{gray!45}3.9\% & \cellcolor{gray!45}4.0\% & \cellcolor{gray!45}3.4\% & \cellcolor{gray!45}0.2\%  \\ 
 \end{tabular}
 \caption{All UMAP datasets, mean of multiclass sensitivities, 0\% training contamination.} 
 \label{tab:multiclass_all_means_umap_0} 
\end{table}
\begin{table} 
 \center 
 \begin{tabular}[h]{c c c c c c c c } 
  measure & \rotatebox{90}{AUC} & \rotatebox{90}{AUC$_w$} & \rotatebox{90}{AUC@0.05} & \rotatebox{90}{precision@0.05} & \rotatebox{90}{TPR@0.05} & \rotatebox{90}{F1@0.05} & \rotatebox{90}{CVOL@0.05}  \\ 
  \hline 
  AUC & \cellcolor{gray!45}0.8\% & \cellcolor{gray!45}2.4\% & 6.3\% & 5.8\% & 4.8\% & 8.2\% & 3.3\%  \\ 
  AUC$_w$ & \cellcolor{gray!30}0.9\% & \cellcolor{gray!15}2.5\% & 6.2\% & 5.8\% & \cellcolor{gray!15}4.6\% & 8.2\% & 3.3\%  \\ 
  AUC@0.05 & 0.9\% & \cellcolor{gray!30}2.5\% & \cellcolor{gray!45}5.0\% & \cellcolor{gray!15}5.4\% & \cellcolor{gray!30}4.0\% & \cellcolor{gray!30}5.9\% & 1.4\%  \\ 
  precision@0.05 & 0.9\% & 2.7\% & \cellcolor{gray!15}6.0\% & \cellcolor{gray!45}4.3\% & 4.9\% & 6.8\% & \cellcolor{gray!30}1.0\%  \\ 
  TPR@0.05 & \cellcolor{gray!15}0.9\% & 2.9\% & \cellcolor{gray!30}5.7\% & \cellcolor{gray!30}5.2\% & \cellcolor{gray!45}3.7\% & 7.1\% & 2.1\%  \\ 
  F1@0.05 & 2.5\% & 10.2\% & 14.2\% & 14.6\% & 10.6\% & \cellcolor{gray!45}4.0\% & \cellcolor{gray!15}1.2\%  \\ 
  CVOL@0.05 & 2.3\% & 9.0\% & 13.8\% & 15.2\% & 11.6\% & \cellcolor{gray!15}6.6\% & \cellcolor{gray!45}0.4\%  \\ 
 \end{tabular}
 \caption{All full datasets, mean of multiclass sensitivities, 0\% training contamination.} 
 \label{tab:multiclass_all_means_full_0} 
\end{table}
Average results from all multi-class datasets scaled down to two dimensions using UMAP are shown in Table~\ref{tab:multiclass_all_means_umap_0}. As can be expected, the volume of the decision region is the most stable selection measure, as it is independent of anomaly class and well represents the case when we want to be conservative, i.e. we do not know anything about anomalies we want to detect. Unfortunately, this criterion is difficult to estimate in high dimensions, as can be seen in Table~\ref{tab:multiclass_all_means_full_0} showing the same quantities measured on datasets with full dimension. We have also found that using our current approach to estimate the volume is not very stable when the training/validation normal data are contaminated by anomalies.

\paragraph{Conclusion:}
Using supervised measures to evaluate anomaly detection is a double-edged sword, as it implies the assumption that the user has examples of anomalies with the same statistical properties as he expects during application. We may wonder if this is a sound assumption. If the user knows something about the nature of anomalies, they should use it to better tune their detector and relying on one or few--shot learning might be better. From this point of view, anomaly detection seems to be difficult to evaluate. A possible solution might be a conservative measure like volume of the decision region, but at the moment, this is difficult to estimate in high dimensions.

\section{Conclusion}
Motivated by practitioners, we have tried to answer the following question: what is the right measure to evaluate anomaly detectors? We have split the investigation into two parts.

At first, we assumed that the practitioner has enough representative samples of anomalies they wish to detect. In this case, our experimental results indicate that the popular AUC seems to be the worst measure, as it is poorly correlated with measures of interest of practitioners, such as \preca\ and \tpra. 
A well better alternative seems to be \aucw\ and AUC@0.05, which put more emphasis on the area of low positive rates. Albeit AUC@0.05 is arbitrary, our study utilizing a large number of datasets indicates that it is more robust to estimation noise than the theoretically more correct \aucw. Also, the 5\% false positive rate threshold is of some practical interest to other researchers~\citep{cogranne2019alaska}.

Secondly, we have investigated an alternative scenario when representative samples of anomalies are not available. In this scenario, the volume of the decision region seems to be a good measure, as it can prevent catastrophic failures. But presently, it is difficult to estimate and it seems to be sensitive to contamination of normal samples by anomalous, which will almost certainly happen in real applications.

Our study clearly demonstrates that despite extensive research in anomaly detection, the correct path of its evaluation is still unclear as it does not adequately reflect their real application deployment. We typically behave like we have representative examples of anomalous samples in the validation data, but in this case, it might be more sensible to split them and perform few--shot learning. Alternatively, if we do not have representative examples, we should evaluate the quality using the volume of the decision region of normal samples which is well theoretically supported but might be difficult to use in practice. 

\begin{acknowledgements}
Research presented in this work has been supported by the Grant SGS18/188/OHK4/3T/14 provided by Czech Technical University in Prague and by the GA\v{C}R project 18-21409S. Also, we would like to thank for the support provided by the RCI institute under CTU.
\end{acknowledgements}

\bibliographystyle{spbasic}
\bibliography{main}

\end{document}